\documentclass[runningheads]{llncs}
\usepackage{graphicx}
\usepackage[vietnamese, romanian, german, english]{babel}
\usepackage{calrsfs}
\usepackage{amsfonts} 
\usepackage{amsmath}
\usepackage{amssymb}
\usepackage{multirow}
\usepackage{makecell}
\usepackage{tabularx}
\usepackage{xcolor}
\usepackage[utf8]{inputenc}
\usepackage{subfig}
\usepackage{tabu}
\usepackage{url}

\DeclareMathAlphabet{\pazocal}{OMS}{zplm}{m}{n}
\DeclareMathOperator*{\mean}{mean}

\newcommand{\Lb}{\pazocal{L}}
\newcommand{\Da}{\pazocal{D}}

\begin{document}
\title{Quantum Statistics-Inspired Neural Attention}
%
%
\author{Aristotelis Charalampous \and
Sotirios Chatzis 
}

\authorrunning{A. Charalampous et al.}

\institute{Cyprus University of Technology, Limassol 3066, Cyprus\\
\email{\{aa.charalampous@edu., sotirios.chatzis@\}cut.ac.cy}
}

\maketitle              
%


\begin{abstract}
Sequence-to-sequence (\emph{encoder-decoder}) models with attention constitute a cornerstone of deep learning research, as they have enabled unprecedented sequential data modeling capabilities. This largely stems from their capacity to infer salient temporal dynamics over long horizons; these are encoded into the obtained neural attention (NA) distributions. However, existing NA formulations essentially constitute point-wise selection mechanisms over the observed source sequences. 
Unfortunately, this assumption fails to account for higher-order dependencies which might be prevalent in real-world data. This paper addresses these limitations by leveraging Quantum-Statistical modeling arguments. Specifically, our work \emph{broadens the notion of NA}, by attempting to account for the case that the NA model becomes inherently incapable of discerning between \emph{individual} source elements; this is assumed to apply due to higher-order temporal dynamics. On the contrary, we postulate that in some cases selection may be feasible only at the level of \emph{pairs} of source sequence elements. To this end, we cast NA into inference of an \emph{attention density matrix (ADM)} approximation. We derive effective training and inference algorithms, and evaluate our approach in the context of a machine translation (MT) application. We perform experiments with challenging benchmark datasets. As we show, our approach yields favorable outcomes in terms of several evaluation metrics.
\end{abstract}
\section{Introduction} \label{introduction}

The key contribution in advancing sequence-to-sequence (encoder-decoder) models accounted to neural attention (NA) networks has rendered the latter a core part of any deep learning toolkit. Indeed, their obtained state-of-the-art results in MT \cite{Bahdanau2014,Luong2015}, abstractive document summarization (ADS) \cite{Manning2017}, speech recognition (SR) \cite{Chorowski2015,Chan2015} and question answering (QA) \cite{Sukhbaatar2015}, amongst others, vouch for their broad applicability. NA network functionality consists in enabling the output layer of sequence-to-sequence models to access a variable-length memory \cite{weston2014,Graves2014,Graves2016} of stored source sequence encodings. Specifically, NA networks infer a time-varying representation of the memory content, which encapsulates the affinity of the stored source sequence encodings with the (elements in the) decoded target sequence \cite{Bahdanau2014}. This is usually referred to as the inferred \emph{context vector representation}, $\boldsymbol{c}_{t}$, of the stored source sequence encodings, and is used to drive output generation (decoding) at time $t\in{1,\dots,T}$. Consequentially, application of NA mechanisms enables encoder-decoder models to perform inference over long temporal horizons; this is not feasible without NA, due to the model having to encode the entire source information into a single fixed vector $\boldsymbol{h}$.
 
This major breakthrough has rendered NA a focal point in recent research developments, mainly revolving around the concept of \emph{Soft-Attention} (SA) \cite{Xu2015}. Despite their wide success, SA mechanisms also suffer from inherent weaknesses, which are actively scrutinized by the research community. One major weakness is that SA essentially implements a soft-selection mechanism over the source sequence encodings. This procedure relies on the assumption of full independence between the selections at each decoding step, and over all the available encodings. Unfortunately, under this simplistic selection rationale, the model is discouraged from exploring the dynamics in the memory space to their full potential.

Structural attention models have been recently proposed as a means of overcoming these issues \cite{wang2017translating,logeswaran2016sentence}. They consist in establishing an SA mechanism which infers a joint attention distribution over consecutive decoding steps. This gives rise to a structural dependencies prior, meant to allow for the model to attend to partial segmentations e.g., pertaining to phrases or other coherent syntactic structures in natural language. However, while this approach may be helpful, for instance, in reducing grammatical and syntax errors in translation outputs, it requires relying on dynamic programming to infer the attention distributions. Unfortunately, such algorithms generally impose exponential-scale complexity. On the other hand, adoption of simplistic relaxations for the sake of computational tractability, e.g. of first-order Markovian dynamics assumptions, or of a maximum length $L$ for each attended segment \cite{wang2017translating,logeswaran2016sentence}, is clearly limiting.

This lack of the existing NA paradigm is the driving force of our work. We posit that enabling the attention mechanism to capture structural ambiguity due to higher-order dynamics in the source element selection process will allow for improved inferential capabilities. To effect this goal, for the first time in the literature, we draw inspiration from quantum statistical concepts.
Quantum statistical systems replace the concept of Categorical distributions with the concept of \emph{Density matrices} \cite{sato2009quantum,warmuth2010bayesian,sordoni2013modeling,chatzis2012quantum}. A density matrix is a square matrix, the order of which corresponds to the number of available system selections; these are referred to as "pure states" in the literature of quantum statistics. Its diagonal elements represent the probabilities of the underlying pure states (which sum to one). In NA terms, this corresponds to a traditional soft selection over the source sequence elements. On the other hand, its non-diagonal elements correspond to the possibility that the system becomes inherently incapable of discerning between individual (pure) states. 
This quantum-statistical modeling rationale does not reinstate the induced higher-order dependencies modeling as introduction of additional latent variables, nor stochastically as joint probabilities \cite{sordoni2013modeling}, similar to related works on robotics \cite{chatzis2012quantum} and information retrieval tasks \cite{sordoni2013modeling}.

To exhibit the efficacy of our novel approach, we implement it into the context of encoder-decoder models with attention. We dub our approach \emph{attention density matrix (ADM) seq2seq networks}. We perform an extensive experimental evaluation of our approach using several benchmark MT datasets. As we show, our method compares favorably to alternative NA mechanisms. 

The remainder of this paper is organized as follows. Section \ref{methodological_background} covers the methodological background. We introduce our novel quantum statistics-inspired regard towards NA in Section \ref{proposed_approach}. In Section \ref{experimental_eval}, we perform the experimental evaluation of our approach. We further discuss and conclude this work in Section \ref{discuss}.

\section{SA for Encoder-Decoder Networks} \label{methodological_background}

Encoder-decoder networks with attention \cite{Bahdanau2014,Luong2015} are formulated as follows: A sequence of source tokens $\mathbf{x} = {x}_{i},\dots,{x}_{T_x}$ are fed to an encoder function ${f}_{enc}$ that generates a sequence of corresponding representations (encodings) $\mathbf{h} = {\boldsymbol{h}}_{1},\dots,{\boldsymbol{h}}_{T_x}$. Typical encoder-decoder models adopt a bidirectional RNN encoder. This yields encodings obtained as the concatenation of the states of the component backward and forward RNNs; that is $\boldsymbol{h}_{i} = [\overrightarrow{\boldsymbol{h}}_{i};\overleftarrow{\boldsymbol{h}}_{i}] \in \mathbb{R}^{D_h}$. On the other hand, the decoder receives at each step, $i$, the current state of the decoder RNN, ${\boldsymbol{s}}_{i}$, and a context vector, ${\boldsymbol{c}}_{i}$, to obtain a distribution over the next element in the target sequence, $ \mathbf{y} = {y}_{1}, \dots, {y}_{{T}_{y}}$.


More formally, the distribution of the generated (decoded) sequence yields:
\begin{equation}
P(\mathbf{y}|\mathbf{x}) = \prod_{i=1}^{T_y}P({y}_{i}|\mathbf{y}_{<i}, \mathbf{x}), \;\;\; P({y}_{i}|\mathbf{y}_{<i}, \mathbf{x}) = \mathrm{softmax}(\mathrm{Dense}({\boldsymbol{s}}_{i}, {\boldsymbol{c}}_{i})) \label{decoding}
\end{equation}

\noindent In these expressions, ${y}_{i}$ is the target sequence element generated by the decoder at step $i$, and $\boldsymbol{s}_i$ is the state of the decoder RNN, $\mathrm{f_{dec}}$, at step $i$:
\begin{equation}
\boldsymbol{s}_i=\mathrm{f_{dec}}(\boldsymbol{s}_{i-1}, y_{i-1}) \in \mathbb{R}^{D_s},
\end{equation}
Typically, the used RNN is a Long Short-Term Memory (LSTM) \cite{Hochreiter1997} or a Gated Recurrent Unit (GRU) network \cite{Cho2014}. On the other hand, the \emph{context vector}, $\boldsymbol{c}_i$, is calculated as a weighted average of the encoder outputs pertaining to the source sequence, $\boldsymbol{h}$ (encodings):
\begin{equation}
\boldsymbol{c}_{i}=\sum_{j=1}^{T_{x}}a_{ij}\boldsymbol{h}_{j} \; \; \; \mathrm{s.t.:} \, \boldsymbol{h}_{j} = \mathrm{f_{enc}}(\boldsymbol{h}_{j-1}, \boldsymbol{x}_{j}) \label{base_context_vector}
\end{equation}

In computing the context vector, $\boldsymbol{c}_{i}$, the used weights $a_{ij}$ assigned to each encoding, $\boldsymbol{h}_{j}$, define a Categorical distribution over the source sequence elements. Essentially, these weights imply establishment of a soft-selection mechanism; the probability of each source element being selected (attended to) is considered analogous to the affinity of the corresponding encoding with the current state of the decoder.
Specifically, we have 
\begin{equation} \label{pure_states_computation}
a_{ij} = \mathrm{softmax}(\alpha(\boldsymbol{s}_{i}, \boldsymbol{h}_{j})), \; \; \mathrm{s.t.:} \sum_{j}{a_{ij}}=1
\end{equation} 
Here, ${\alpha}(\cdot)$ is an \emph{alignment network} \cite{Bahdanau2014}. In the most usual case, the alignment network takes the form of an additive variant that reads 
\begin{equation} \label{soft-attention-scores}
\alpha(\boldsymbol{s}_{i}, \boldsymbol{h}_{j}) = \boldsymbol{u}^{\top}_{\alpha}\mathrm{tanh}(\boldsymbol{W}_{\alpha}\boldsymbol{s}_{i} + \boldsymbol{V}_{\alpha}\boldsymbol{h}_j + \boldsymbol{b}_{\alpha}),
\end{equation}
where $\boldsymbol{W}_{\alpha}$ and $\boldsymbol{V}_{\alpha}$ are trainable weight matrices, while $\boldsymbol{u}_{\alpha}$ and $\boldsymbol{b}_{\alpha}$ are trainable parameter vectors of the same size. 

\section{Proposed Approach} \label{proposed_approach}

The formulation of our approach draws inspiration from the concept of the \emph{density matrix}. A density matrix $\boldsymbol{\Phi}$ is symmetric, $\boldsymbol{\Phi} = \boldsymbol{\Phi}^{\top}$, positive-semidefinite, $\boldsymbol{\Phi} \succeq 0$, and of trace one. It constitutes a natural extension of Categorical probability distributions in the context of quantum statistics, whereby a modeled system may become inherently incapable of assigning selection probabilities to individual ("pure") states; assumingly due to underlying higher-order dynamics. 

Indeed, a Categorical distribution can be represented via a density matrix of diagonal form, referring to Gleason's theorem \cite{gleason1957measures}. On the other hand, the non-diagonal elements of a density matrix correspond to the probabilities of the so-called "mixed states". These essentially reflect a situation where the quantum-statistical system "lives" in a mixture of states, as opposed to selecting to live in a specific "pure" state.

Drawing from this inspiration, in this work we suggest that we infer attention dynamics in the form of a \emph{pseudo}-density matrix, $\boldsymbol{\Psi}$. Under our proposed approach, the diagonal elements of $\boldsymbol{\Psi}$ constitute \textit{attention scores} that correspond to the system selecting to attend to one of  
the available source sequence elements, in a fashion similar to the selection principles of SA. On the other hand, the non-diagonal elements of $\boldsymbol{\Psi}$ constitute scores tied to the possibility that selection may not be feasible for the model at the level of individual source sequence elements. On the contrary, the model may only become capable of performing selection at the level of pairs of elements, while lacking the capacity to discern which specifically of the pair elements it must attend to. 

Let us consider an NA mechanism performing attention over $N$ source sequence elements. We succinctly represent the above-prescribed concepts and assumptions by introducing the notion of the Attention Density Matrix (ADM). Considering the decoding step $i$, we define the pseudo-density matrix:
\begin{equation}
\boldsymbol{\Psi}_i\triangleq\left[\begin{array}{cccc}
\alpha_{i\,1} & \sigma_{i\,(1,2)} & \dots & \sigma_{i\,(1,N)}\\
\sigma_{{i\,(2,1)}} & \alpha_{i2} & \dots & \sigma_{{i\,(2,N)}}\\
\vdots & \vdots & \ddots & \vdots\\
\sigma_{{i\,(N, 1)}} & \sigma_{i\,({N, 2)}} & \dots & \alpha_{iN}
\end{array}\right] \label{attention_density_matrix}
\end{equation}
We dub this matrix the ADM of the introduced NA model at time $i$. Its diagonal elements $\alpha_{ij}$ correspond to the attention scores of attending to source element $j$ at decoding step $i$, defined in the classical sense. On the other hand, the non-diagonal elements, $\sigma_{i\,(j,k)} \in \mathbb{R}$, correspond to the score pertaining to the system's inability to discern between source elements $j$ and $k$ at decoding step $i$; instead, it selects to attend to their pair, while being oblivious as to which specific element of the pair is the most appropriate to attend to. In this way, we seek to elevate the system's \emph{uncertainty} as an additional model feature. Apparently, since the ADM $\boldsymbol{\Psi}_i$ is symmetric, it holds $\sigma_{i\,(j,k)}=\sigma_{i\,(k,j)},\;\forall\,j\neq k$. 

Note that the introduced ADM may reduce to the Categorical attention distribution employed in conventional SA, by setting its non-diagonal elements equal to zero. In addition, we emphasize that we elect not to enforce a positive-semidefiniteness constraint on the ADMs, $\boldsymbol{\Psi}_i$, nor do we express their elements as probabilities. This is reasonable, since our approach does not model an actual quantum-statistical system; we draw inspiration from it, in an effort to effect computationally efficient and elegant modeling of attention dynamics.

Let us now proceed to derivation of the context vector representations under our ADM-based NA mechanism. To this end, we begin from the classical definition of context vectors adopted under the SA scheme. From definition (\ref{base_context_vector}), it follows that the expression of SA context vectors can be rewritten in the following form:
\begin{equation}
\boldsymbol{c}_{i}=\sum_{j=1}^{N}a_{ij}\boldsymbol{h}_{j} = \boldsymbol{H} \boldsymbol{\alpha}_{i}\\
\end{equation} \label{base_context_vector_rewritten}
where $\boldsymbol{H}=[\boldsymbol{h}_{j}]_{j=1}^{N}$ is the columnwise concatenation of the vectors set $\{\boldsymbol{h}_{j}\}_{j=1}^{N}$. 

On this basis, we seek to devise an effective means of generalizing the expression (7) in the context of our quantum-statistical formulation. To this end, we perform row-wise averaging of the ADM-induced, state-related quantities. This is an efficient formulation; we defer a supporting discussion of this aspect in Section \ref{discuss}. 

Specifically, our quantum statistics-inspired view towards NA materializes by adopting the same functional form for the derived context vectors, as in (7). This brings to the fore a natural way to generalize attention weight formulation under our approach, with the same output dimensionality. We postulate
\begin{equation}
\boldsymbol{c}_{i}=\boldsymbol{H}\boldsymbol{\omega}_{i},
\; \; \; \mathrm{s.t.:} \, \boldsymbol{\omega}_{i}=\mathrm{softmax}(\;\mean\limits_{1\leq j\leq N}[\boldsymbol{\Psi}_{i}]_{jk}\;)
\label{quantum_context_vector}
\end{equation}
where $[\boldsymbol{\Psi}_{i}]_{jk}$ is the $(j,k)$th element of matrix $\boldsymbol{\Psi}_{i}$.



The key difference from (7) is that we replace the attention weights of conventional SA with the probabilities of the row-wise mean score of state mixtures, defined in (\ref{attention_density_matrix}). The rationale of this statistical function is to roughly describe mean \emph{temporal variation} during source element selection. In essence, we establish a mean-pooling mechanism, which amortizes the variance among scores pertaining to each source sequence element. This is in line with recent work related to enhancing sentence embeddings by leveraging mean-pooling operations across multiple NLP tasks, e.g. \cite{chen2018enhancing}.

Having provided the definition of the introduced ADM concept, as well as the corresponding derivations pertaining to context vector inference under our approach, we now elaborate on how we can compute the ADM elements. We begin with the diagonal elements, $\alpha_{ij}$, which correspond to classical soft-selection over the source sequence elements. In this work, we obtain these quantities by simply adopting the \emph{score} functions employed in conventional SA. Indeed, this is a natural selection, since the diagonal elements of the ADM have the same physical interpretation as the scores in conventional SA.

Let us now turn to the non-diagonal elements, $\sigma_{i\,(j,k)}$ of the postulated ADM, $\boldsymbol{\Psi}_i$, at time $i$. To introduce an effective way of computing them, we draw inspiration from their physical interpretation. As we have already discussed, a non-diagonal ADM element, $\sigma_{i\,(j,k)}$, assigns a 
score to the scenario that the postulated NA model cannot effectively discern between the $j$th and the $k$th source sequence encoding, $\{\boldsymbol{h}_j, \boldsymbol{h}_k\}\; \forall{j\neq k}$. Therefore, to obtain these  
scores, we need to introduce a transformation that can effectively reflect two distinct aspects: (i) the affinity between pairs of source sequence encodings, $\{\boldsymbol{h}_j, \boldsymbol{h}_k\}\; \forall{j\neq k}$; and (ii) the affinity of the current decoder state, $\boldsymbol{s}_i$, with the latent dynamics shared by the encodings $\boldsymbol{h}_j$ and  $\boldsymbol{h}_k$. 

To effect this goal, at each decoding step, $i$, we first compute a set of $D$-dimensional score vectors between each of the available source sequence encodings.
We define the 3-dimensional tensor $\boldsymbol{L}=[l_{(j,k)}]_{j,k}$, where: 
\begin{equation} \label{order-1-quantum}
l_{(j,k)}=\mathrm{tanh}(\boldsymbol{h}_{j}+\boldsymbol{h}_{k}),
\end{equation}
This way, we yield a high-dimensional representation of the correlations between all pairs of the available source sequence encodings.
This enables us to further extract rich information regarding the latent dynamics shared between all possible pairs of encodings, and in respect to the current decoder state. Specifically, the sought higher-order interaction dynamics matrix is computed through Eq. (\ref{luong:quantum}) using the tensor $\boldsymbol{L}$ in (\ref{order-1-quantum}). We have
\begin{equation} \label{luong:quantum}
\boldsymbol{M}_{i}= w_{s}\boldsymbol{L}\boldsymbol{s}_{i},
\end{equation}
where $w_{s}\in\mathbb{R}$ is a trainable weight scalar. Because of the multiplicative style it employs, resembling the attention mechanism in \cite{Luong2015}, we refer to it as the \textit{Multiplicative Quantum-Theoretical (MQT)} scheme. We make this distinction as it is possible to follow a different scheme, by replacing Eqs. (\ref{order-1-quantum} - \ref{luong:quantum}) with:

\begin{equation} \label{bahdanau:quantum}
\boldsymbol{M}_{i}= \mathrm{tanh}(\boldsymbol{L} + \boldsymbol{s}_{i}) \boldsymbol{w}_{s},
\end{equation}
where $\boldsymbol{w}_{s}$ is a trainable weights vector.
\noindent We dub this alternative formulation as the \textit{Additive Quantum-Theoretical (AQT)} scheme, as it resembles \cite{Bahdanau2014}.

Eventually, we use the non-diagonal elements of the so-derived matrix $\boldsymbol{M}_{i}$ as the non-diagonal elements of the sought ADM, $\boldsymbol{\Psi}_{i}$: $\sigma_{i(j,k)}=[\boldsymbol{M}_i]_{(j,k)}$. 

\noindent\textbf{Relation to recent work.} As demonstrated above, our approach enables capturing higher-order interactions in the
computation of context vectors. Recent efforts
toward generalizing neural attention, by deriving more complex attention
distributions, contrast our proposed approach. We refer again to \cite{structuredAttention}, which recently
introduced \emph{structured attention}, whereby the attention probabilities are considered to be
interdependent over consecutive time-steps; for instance, they postulate
the first-order Markov dynamics assumption: 
\begin{equation}
p(\{\boldsymbol{z}_{t}\}_{t=1}^{T};\Da)=p(\boldsymbol{z}_{1};\Da)\prod_{t=1}^{T-1}p(\boldsymbol{z}_{t+1}|\boldsymbol{z}_{t};\Da),
\label{eq:Structured}
\end{equation}
where $\{z_{t}^{i}\}_{i=1}^{N}$,
$z_{t}^{i}\in\{0,1\}$, with $z_{t}^{i}=1$ being a set of binary latent variables indicating the source element the model attends to at time $t$. To perform training and inference, the computationally complex forward-backward algorithm \cite{rabiner1989tutorial} has to be used. This also discourages the method's use in scale, contrary to our approach which imposes minimal additional computational costs. This is the case as all the extra computations of our model, described in Eqs. (10)-(12), constitute single additional forward propagation computations; this renders our model comparable to classical SA in terms of computational costs.

\noindent\textbf{Inference Algorithm.} To perform target decoding by means of a \emph{seq2seq} model with ADM, we resort to \emph{Beam sampling} \cite{russell2016artificial}. In our experiments, \emph{Beam} \emph{width} is set to ten across all MT tasks. 

\noindent\textbf{Training Algorithm.} To perform training of a \emph{seq2seq} model with ADM, we rely on minimization of an appropriate loss function, $\Lb(\theta)$, where $\theta$ is the set of model parameters, in a similar fashion to conventional models. For example, to address an MT task, one may use the categorical cross-entropy between the generated decoding probabilities (\ref{decoding}) and the available groundtruth.





\section{Experimental Evaluation} \label{experimental_eval}

\begin{table*}[t]
\caption{Translation results on the En$\leftrightarrow$Vi and En$\leftrightarrow$Ro pairs.}
\small
\centering{}%
\begin{tabu}{|c|c|c|c|c|[1.5pt]c|c|c|c|}
\hline 
\multirow{3}{*}{Method} & \multicolumn{8}{c|}{BLEU}\tabularnewline
& \multicolumn{2}{c}{En$\rightarrow$Vi} & \multicolumn{2}{c}{Vi$\rightarrow$En} & \multicolumn{2}{c}{En$\rightarrow$Ro} & \multicolumn{2}{c|}{Ro$\rightarrow$En}
\tabularnewline
\cline{2-9} 
 & dev & test & dev & test & dev & test & dev & test\tabularnewline
\hline
\multirow{1}{*}{Baseline} & \multirow{1}{*}{22.90} & \multirow{1}{*}{24.11} & \multirow{1}{*}{19.14} & \multirow{1}{*}{20.76} & \multirow{1}{*}{13.65} & \multirow{1}{*}{15.31} & \multirow{1}{*}{17.15} & \multirow{1}{*}{16.74}
\tabularnewline
\cline{2-9}
\multirow{1}{*}{\makecell{Structured Attention}} & \multirow{1}{*}{16.81} & \multirow{1}{*}{17.00} & \multirow{1}{*}{17.19} & \multirow{1}{*}{18.08} & \multirow{1}{*}{7.04} & \multirow{1}{*}{7.03} & \multirow{1}{*}{11.02} & \multirow{1}{*}{11.08}
\tabularnewline
\cline{2-9}
\multirow{1}{*}{\textbf{MQT}} & \multirow{1}{*}{23.41} & \multirow{1}{*}{25.34} & \multirow{1}{*}{20.65} & \multirow{1}{*}{22.23} & \multirow{1}{*}{15.01} & \multirow{1}{*}{16.95} & \multirow{1}{*}{19.03} & \multirow{1}{*}{19.09}
\tabularnewline
\cline{2-9}
\multirow{1}{*}{AQT} & \multirow{1}{*}{22.71} & \multirow{1}{*}{24.34} & \multirow{1}{*}{19.95} & \multirow{1}{*}{21.33} & \multirow{1}{*}{3.98}
& \multirow{1}{*}{4.56} & \multirow{1}{*}{5.08} & \multirow{1}{*}{5.62}
\tabularnewline
\hline
\end{tabu}
\label{translation:results}
\end{table*}


Our experiments make use of publicly available corpora, namely IWSLT'15 English to and from Vietnamese (IWSLT En$\leftrightarrow$Vi) and 
WMT'16 English to and from Romanian (WMT En$\leftrightarrow$Ro). 
We benchmark our models against word-based vocabularies and present our results in terms of the BLEU score \cite{blue}. Note that, in this paper, we employ relatively small datasets to better demonstrate the applicability and generalization strength of our approach. To allow seamless handling of rare words, we utilize \emph{byte pair encoding} (BPE) \cite{sennrich2015neural} for En$\leftrightarrow$Ro.

\subsection{Datasets}

\noindent\textbf{IWSLT En$\leftrightarrow$Vi.} One benchmark we consider is the IWSLT En$\leftrightarrow$Vi machine translation task.
The dataset includes $\sim$133K training sentence pairs, from translated TED talks,
provided by the IWSLT 2015 Evaluation Campaign \cite{cettolo2015iwslt}. Following the same preprocessing steps in \cite{Luong2015,raffel2017online}, we use the TED tst2012 ($\sim$1.5K sentences) as a validation set for hyperparameter tuning and TED tst2013 ($\sim$1.2K sentences) as a test set. The Vietnamese and English vocabulary sizes are $\sim$7.7K and $\sim$17.2K respectively.


\noindent\textbf{WMT En$\leftrightarrow$Ro.} A second dataset our evaluation addresses is the WMT En$\leftrightarrow$Ro machine translation task. The dataset consists of $\sim$400K parallel sentences. The shared vocabulary sizes (obtained from BPE with 32K operations) total $\sim$31K words.  We use newsdev2016 as our development set, and newstest2016 as our test set (both with 1.9K sentences).

\subsection{Experimental setup} 

\begin{table}[t]
\small
\caption{Vi$\rightarrow$En, dev set - Examples (a) 281 and (b) 1086.}
\begin{tabularx}{\textwidth}{|X|} \hline
\textbf{Reference Translation}\tabularnewline \hline 
(a) And for me that means spending time thinking and talking about the poor, the disadvantaged, those who will never get to TED. \\
(b) Now the transition point happened when these communities got so close that, in fact, they got together and decided to write down the whole recipe for the community together on one string of DNA.
\tabularnewline \hline \hline
\textbf{Generated Translation - Baseline}\tabularnewline \hline
(a) For me, it means \textcolor{orange}{to be} spending time to think about the poor, who have a situation, about who don't have a chance to \textbf{[verb]} TED. \\
(b) The time that transitions \textbf{[verb]} when communities are closer together, in  \textcolor{blue}{reality}, they're together, form a  \textcolor{red}{collective tool} for the whole  \textcolor{red}{chromosome} on a single  \textcolor{blue}{array} of DNA.
\tabularnewline \hline \hline
\textbf{Generated Translation - Structured Attention}\tabularnewline \hline
(a) For me, that means for time to think, \textcolor{orange}{to think} about the poor, to talk about people who are not \textcolor{orange}{having to} go to TED. \\
(b) The time \textcolor{red}{$<$unk$>$} occurs when the communities are close to each other, in fact\textcolor{orange}{, in fact,} they're together to form a shared structure for the entire \textcolor{red}{body}.
\tabularnewline \hline \hline
\textbf{Generated Translation - MQT}\tabularnewline
\hline 
(a) For me, it means to spend time to think, to talk about the poor people, who have a \textcolor{blue}{difficult situation}, about who there is \textcolor{blue}{no opportunity} to go to TED. \\
(b) A time of \textcolor{red}{communication} happens when communities are close to each other, in fact, they together form a \textcolor{blue}{formula} that is shared for the whole \textcolor{blue}{colony} on a single \textcolor{blue}{array} of DNA.
\tabularnewline \hline
\end{tabularx}
\label{vi-en:translations}
\end{table}

We perform dropout regularization of the trained models, with a dropout rate equal to 0.2. We minimize $\Lb(\phi)$ by employing the Adam \cite{adam} optimizer with its default settings for En$\leftrightarrow$Vi and simple stochastic gradient descent (SGD) for En$\leftrightarrow$Ro\footnote{En$\leftrightarrow$Vi models are trained for $\sim$12 epochs; En$\leftrightarrow$Ro for$\sim$12 epochs for structured attention models and $\sim$4 for the rest.}. We preserve homogeneity throughout the trained architectures as follows.
Both the encoders and the decoders of all the evaluated models are presented with 256-dimensional \emph{trainable} word embeddings. The maximum inference length is set to 50. We utilize 2-layer BiLSTM encoders, and 2-layer LSTM decoders; all comprise 256-dimensional hidden states on each layer. For the remainder of hyper-parameters, we adopt the default settings used in the code\footnote{https://github.com/harvardnlp/struct-attn.} provided by the authors in \cite{structuredAttention} for structured attention models and the code in \cite{luong17} for the rest. Except specified otherwise, the default settings used by the latter for En$\leftrightarrow$Vi also apply to En$\leftrightarrow$Ro.



\subsection{Results} 
\begin{table}[t]
\caption{Ro$\rightarrow$En, dev set - Examples (a) 5 and (b) 182.}
\begin{tabularx}{\textwidth} {|X|} \hline 
\textbf{Reference Translation}\tabularnewline \hline 
(a) Dirceu is the most senior member of the ruling Workers' Party to be taken into custody in connection with the scheme.
\\
(b) With one voice the lobbyists talked about a hoped-for ability in Turnbull to make the public argument, to cut the political deal and get tough things done.
\tabularnewline \hline \hline
\textbf{Generated Translation - Baseline}\tabularnewline \hline
(a) He is the oldest member of the \textcolor{orange}{Dutch People's Party} on \textcolor{orange}{Human Rights} in custody for \textcolor{orange}{the} \textcolor{blue}{links} with this scheme.
\\
(b) The representatives of \textcolor{blue}{lobbyists} have spoken about their hope in the ability of \textcolor{orange}{Turngl} to \textcolor{blue}{satisfy} the public interest, to reach a political agreement and to do things well.
\tabularnewline \hline \hline
\textbf{Generated Translation - Structured Attention}\tabularnewline \hline
(a) \textcolor{orange}{It} is the oldest member of the \textcolor{orange}{Mandi} \textcolor{orange}{of the Massi} in \textcolor{orange}{the} government \textcolor{red}{in the government}.
\\
(b) The representatives of the \textcolor{blue}{interest groups} have spoken \textcolor{red}{in mind} about their hope to \textcolor{blue}{meet} the public interest, to achieve a political and good thing.
\tabularnewline \hline
\textbf{Generated Translation - MQT}\tabularnewline \hline 
(a) \textcolor{orange}{Dirre} is the oldest member of the \textcolor{orange}{People's Party} in government \textcolor{blue}{held} in custody for \textcolor{blue}{ties} with this scheme.
\\
(b) The representatives of \textcolor{blue}{interest groups} have spoken \textcolor{orange}{to} \textcolor{blue}{unison} about their hope in \textcolor{orange}{Turkey's} ability to \textcolor{blue}{meet} the public interest, to reach a political agreement and to do things well.
\tabularnewline \hline
\end{tabularx}
\label{ro-en:translations}
\end{table}

Table \ref{translation:results} shows superior performance for our \emph{multiplicative} approach. In addition, note that despite their extended training requirements, structured attention models demonstrate an inability to properly capture long-temporal information, both score and output-wise, as presented in Tables \ref{vi-en:translations} and \ref{ro-en:translations}. These showcase some characteristic examples of generated translations for a hands-on inspection of model outputs. We annotate deviations from the reference translation with orange and red, for minor
and major deviations
respectively. Synonyms are highlighted with blue. We also indicate missing tokens, such as verbs, articles and adjectives, by adding the \textbf{[$<$token$>$]} identifier.

\section{Discussion} \label{discuss}

In this section, we want to further explore how and why our proposed approach can enable
better utilization of infrequent words through second-order interactions. Furthermore, we outline technical augmentations that we consider as future research directives.

\subsection{Model uncertainty} 

\begin{table*}[t]
\caption{Rare word mean reference frequency deviation}
\small
\centering{}%
\begin{tabu}{|c|c|c|c|c|}
\hline 
\multirow{3}{*}{Language Pair} &  & \multicolumn{3}{c|}{Deviation (\%)}\tabularnewline
\cline{3-5} 
 & \makecell{Mean reference \\ frequency (\%)} & Baseline & \makecell{Structured \\ Attention} & MQT\tabularnewline
\hline
\multirow{1}{*}{En$\rightarrow$Vi} & \multirow{1}{*}{3.59} & \multirow{1}{*}{-4.82} & \multirow{1}{*}{11.57} & \multirow{1}{*}{\textit{\textbf{0.28}}}
\tabularnewline
\cline{2-5}
\multirow{1}{*}{\makecell{Vi$\rightarrow$En}} & \multirow{1}{*}{8.91} & \multirow{1}{*}{-10.77} & \multirow{1}{*}{-20.85} & \multirow{1}{*}{\textit{\textbf{-8.68}}}
\tabularnewline
\cline{2-5}
\multirow{1}{*}{En$\rightarrow$Ro} & \multirow{1}{*}{7.00} & \multirow{1}{*}{\textbf{4.59}} & \multirow{1}{*}{\textit{14.61}} & \multirow{1}{*}{-7.87}
\tabularnewline
\cline{2-5}
\multirow{1}{*}{Ro$\rightarrow$En} & \multirow{1}{*}{6.75} & \multirow{1}{*}{34.23} & \multirow{1}{*}{37.47} & \multirow{1}{*}{\textbf{\textit{23.17}}}
\tabularnewline
\hline
\end{tabu}
\label{rare-words-table}
\end{table*}


A first case study is inspired from the interesting work of \cite{ott2018analyzing}. Therein, the major claim is that if model and data distributions match, then samples drawn from both should also match. As a broader extension to our evaluation, we present deviation from unigram word distributions (across 1 - 30\% frequency groups). Table \ref{rare-words-table} reveals how well our evaluated models estimate said frequencies. Words are split into groups based on their appearance in their respective training datasets. We present this against development set frequencies. Note that evaluation criteria include not only \textit{deviation} but also \textit{least over-representation} (favored to any magnitude of under-representation); these are presented in bold and italics, respectively. This is because swapping a frequent for a rare word would not be as harmful as the reverse, in terms of translation quality
In 3 out of 4 cases, our approach achieves close representation of target frequencies.



\subsection{High rank matrix approximation}
The capacity of most natural language models is crippled by their inability to cope in highly context-dependent settings \cite{yang2017breaking}. Admittedly, this shortcoming is due to their limited capacity to capture complex hierarchical dependencies. To address this issue, we need to devise computationally efficient ways of capturing higher-order dynamics. A first step towards this goal is offered by our approach. However, our approach is limited to second-order interactions; in addition, to allow for computational efficiency, we have resorted to a mean-pooling solution in the computation of the final context vector.
A more general solution would be to resort to spectral decomposition, which would require performing eigen/tensor decomposition.
However, differentiating this operation during back-propagation may lead to numerical instability issues \cite{dang2018eigendecomposition}, rendering it non-differentiable. 
We aim to examine solutions to these issues in our future work.
\section{Conclusions}

In this work, we introduced a novel regard towards formulating attention layers in \emph{seq2seq}-type models. Our work was inspired from the quest of a more expressive way of computing dependencies between input and output sequences. Specifically, our aim was to enable capturing of second-order dependencies between the source sequence encodings and the generated output sequences. 

To effect this goal, for the first time in the literature, we leveraged concepts from the field of quantum statistics. We cast the operation of the attention layer into the computation of the \emph{Attention Density Matrix}, which expresses how pairs of source sequence elements correlated with each other, and jointly with the generated output sequence. Our formulation of the ADM was based on \emph{Density Matrix} theory; it is an attempt to encapsulate the core concepts of the Density Matrix in the context of attention networks, without adhering though to the exact definition and properties of density matrices.

We exhibited the merits of our approach on \emph{seq2seq} architectures addressing competitive MT tasks.
We have showed that the unique modeling capacity of our approach translates into better handling of (rare) words in the model outputs. Hence, this finding offers a quite plausible explanation of the obtained improvement in the achieved BLEU scores.
Finally, we emphasize inference using our method entails minor computational overhead compared to conventional SA, with only a single extra forward-propagation computation.

\bibliography{quantum_pakdd}
\bibliographystyle{splncs04}
\end{document}